\documentclass[letterpaper,10pt,twocolumn]{article}
\usepackage[dvips]{graphicx}

\title{Using state space differential geometry \\ for nonlinear blind source separation}
\author{David N. Levin \\ University of Chicago\\ d-levin@uchicago.edu \\ \\ This paper is posted online at http://arxiv.org/abs/cs/0612096}
\date{\today}

\begin{document}

\maketitle

\begin{abstract}
\small{Given a time series of multicomponent measurements of an evolving stimulus, nonlinear blind source separation (BSS) seeks to find a ``source" time series, comprised of statistically independent combinations of the measured components. In this paper, we seek a source time series with \textit{local} velocity cross correlations that vanish \textit{everywhere in stimulus state space}. However, in an earlier paper the local velocity correlation matrix was shown to constitute a metric on state space. Therefore, nonlinear BSS  maps onto a problem of differential geometry: given the metric observed in the measurement coordinate system, find another coordinate system in which the metric is diagonal everywhere. We show how to determine if the observed data are separable in this way, and, if they are, we show how to construct the required transformation to the source coordinate system, which is essentially unique except for an unknown rotation that can be found by applying the methods of linear BSS. Thus, the proposed technique solves nonlinear BSS in many situations or, at least, reduces it to linear BSS, without the use of probabilistic, parametric, or iterative procedures. This paper also describes a generalization of this methodology that performs nonlinear independent subspace separation.  In every case, the resulting decomposition of the observed data is an intrinsic property of the stimulus' evolution in the sense that it does not depend on the way the observer chooses to view it (e.g., the choice of the observing machine's sensors). In other words, the decomposition is a property of the evolution of the ``real" stimulus that is ``out there" broadcasting energy to the observer.  The technique is illustrated with analytic and numerical examples.}
\end{abstract}

\section*{\begin{center} 1. INTRODUCTION \end{center}}

Humans can often decompose a signal time series into components from different sources and then use the results to construct a representation of the evolving state of each source.  For example, suppose someone is listening to a monaural audio track of a ``cocktail'' party that records the utterances of a speaker in the presence of other sound sources (``noise'').  Knowing little or nothing about the sound sources, most listeners can extract the speech content of such a recording for a large range of signal-to-noise ratios and for many different types of speakers and noise processes.  Remarkably, the brain does this by processing the sensory output of the ear, which is \textit{nonlinearly} related to the superposed acoustic waves from the sound sources. The success of this natural biological ``experiment'' has inspired many attempts to devise numerical algorithms that perform this type of ``blind'' source separation (BSS).

Most efforts have focused on the problem of separating signals that are linearly mixed.  For example, suppose that $\tilde{x}(t)$ is a multiplet of $n$ observed time-dependent signals ($\tilde{x}_k \mbox{ for } k = 1,2, \ldots ,n$), and suppose that $\tilde{x}$ is a linear combination of $n$ source signals $x_k$
\begin{equation}
\tilde{x}(t) = M \, x(t)
\end{equation}
where $M$ is an unknown $n \, \mbox{x} \, n$ matrix. If the components of $x$ are assumed to be statistically independent of one another, we can attempt to compute $M$ by imposing this constraint.  For example, second-order statistical independence can be imposed by requiring that cross-correlations between $x$ components vanish; i.e., we can demand that $<xx>$ be a diagonal matrix.  This determines $M$ up to $n \mbox{-dimensional}$ rotations, scaling transformations, and subspace permutations. In most cases, the unknown rotation can be computed by imposing some criterion of higher-order (greater than second-order) statistical independence, and many statistical objective functions have been devised for this purpose~\cite{Hyvarinen}. Finally, it is worth mentioning that there is a related linear BSS problem, sometimes called independent subspace analysis, in which the source components can be partitioned into groups, so that components from different groups are statistically independent but components belonging to the same group may be dependent~\cite{Nishimori}. This weaker assumption is utilized by imposing less restrictive criteria of statistical independence; e.g., by demanding that $<xx>$  be block-diagonal, instead of fully diagonal. Reference~\cite{Hyvarinen} describes many fruitful investigations of the linear BSS problem, as well as a host of applications.

In the nonlinear BSS problem, we consider a time series of observations $\tilde{x}(t)$ that are instantaneous mixtures of source components $x(t)$
\begin{equation}
 \tilde{x}(t) = f[x(t)]
\end{equation}
where $f$ is an unknown, possibly nonlinear, $n \mbox{-component}$ mixing function. As before, the objective is to compute the mixing function from the observed values of $\tilde{x}$ and the statistical independence of the components of $x$. In many approaches to this problem, the mapping $f$ is parametrically modeled by a neural network. The network's weights can be computed by maximizing the statistical independence of its input, as measured by mutual information or other criteria~\cite{Yang}. Alternatively, the network's parameters can be determined by probabilistic learning methods~\cite{Haykin}, although computational expense can be great and results may be degraded if the system is attracted to a local minimum. Other investigators~\cite{Lee} have employed collections of linear BSS algorithms to exploit statistical information in clustered subsets of the observed data, which have been identified with nonlinear techniques. Furthermore, significant attention has been directed at the separation of post-nonlinear mixtures, a special case of nonlinear BSS in which each observed signal component is a nonlinear function of a fixed linear combination of source components~\cite{Taleb}. For a review of these and many other approaches to nonlinear BSS, see Ref.~\cite{Jutten}.

In nonlinear BSS, the search space of all smooth real functions is much larger than the search space in linear BSS (i.e., the space of linear transformations). This suggests that nonlinear BSS will require the imposition of much stronger criteria for statistical independence than were employed to solve the linear problem. Instead of using higher-order statistical constraints for this purpose, we assume that the components of the source's time derivatives $\dot{x}$ are statistically independent of one another \textit{in each neighborhood of $x \mbox{-space}$}. This means that all of the \textit{local} cross-correlations of these derivatives must vanish; i.e., $<(\dot{x}_k-\bar{\dot{x}}_k) \, (\dot{x}_l-\bar{\dot{x}}_l)>_{x}$ must be diagonal for each value of $x$, where the bracket denotes the time average over the trajectory's segments in a small neighborhood of $x$ and where $\bar{\dot{x}} = <\dot{x}>_x$, the local time average of $\dot{x}$. Given the observed data $\tilde{x}(t)$, each of these conditions is a constraint on $f^{-1}$. The technical problem is to simultaneously impose all of these constraints, which are infinite in number. Notice that $\tilde{x}(t)$ and $x(t)$ represent the same stimulus trajectory in two different coordinate systems on the stimulus state space. So, the challenge is two-fold: 1) we must use the observed stimulus trajectory in the $\tilde{x}$ coordinate system in order to determine if there is another coordinate system ($x$) in which the local velocity correlation matrix is diagonal everywhere; 2) if such a coordinate system exists, we must find the transformation ($f^{-1}$) to it. This can be done in the framework of an earlier paper~\cite{Levin}, in which the velocity correlation matrix was shown to constitute a metric on the stimulus state space. In this context, nonlinear BSS  maps onto the following problem of differential geometry~\cite{Weinberg}: given the metric observed in the $\tilde{x}$ coordinate system, ascertain if there is another coordinate system $x$ in which the metric is diagonal, and, if such a coordinate system exists, find the transformation to it. Both of these tasks can be accomplished with the help of the parallel transfer operation and the curvature tensor that the observed trajectory $\tilde{x}(t)$ induces on the stimulus state space. This paper actually describes a more general method that also performs nonlinear independent subspace separation.  Specifically, we show how to systematically determine if there is another ($x$) coordinate system in which the metric is block-diagonalized for all $x$, and, if that transformation exists, we show how to construct it.

There are several ways in which the methodology of this paper differs from previously-described techniques. First of all, the proposed method exploits second-order statistical constraints on source time derivatives that are \textit{locally} defined in the stimulus state space, in contrast to the usual criteria for statistical independence that are \textit{global} conditions on the source time series or its time derivatives~\cite{Lagrange}.  In addition, in this paper, these constraints are solved in a "deterministic" manner, without the need for probabilistic learning methods. Nor have we found it necessary to parameterize the unknown function $f$ with a neural network architecture or other means. Furthermore, unlike many other approaches, higher-order statistical constraints are not used to unravel the nonlinearities of the mixing function, although they may be useful once the problem has been reduced to linear BSS. Finally, the use of differential geometry in this paper should not be confused with existing applications of differential geometry to BSS. In our case, \textit{a metric on the system's state space} is derived from the observed measurement trajectory. The separability constraints, which are relatively easy to derive in the source ($x$) coordinate system, are then formulated as coordinate-system-independent conditions on the space's data-derived curvature tensor. These constraints can be solved in the measurement ($\tilde{x}$) coordinate system in order to determine if the observed process is separable and, if so, to find the transformation to the source coordinate system. In contrast, other authors~\cite{Amari} define \textit{a metric on a completely different space, the search space of possible mixing functions,} so that "natural" (i.e., covariant) differentiation can be used to expedite the search for the function that optimizes the fit to the observed data.

The next section describes the theoretical framework of the new method, and Section 3 describes illustrative examples. Specifically, in Section 3, we show how almost any classical physical description of two non-interacting processes can be used to construct trajectories that are separable by means of the proposed BSS technique. Section 3 also briefly describes a specific numerical ``experiment" in which the new method was used to perform nonlinear independent subspace separation. The implications of this work are discussed in Section 4. A detailed description of the numerical example is given in the Appendix.

\section*{\begin{center} 2. THEORY \end{center}}

This Section shows how to test the stimulus trajectory observed in the measurement-defined coordinate system in order to determine if the data are separable and, if it is, to find the transformation to a source coordinate system in which groups of coordinate components are independent of one another.

Suppose that $x(t)$ ($x_k \mbox{ for } k = 1,2, \ldots ,n$) denotes the multiplet of state space coordinates of an evolving physical stimulus, and let $\tilde{x}(t)$ be an $n \mbox{-dimensional}$ multiplet of measurements that are instantaneous mixtures of the $x_k$, as shown in Eq.(2). The unknown function $f$ is assumed to be real, differentiable, and invertible. In most physical situations, invertibility is a weak assumption for the following reason. Suppose that the sensors of the observing machine produce at least $2n+1$ signals, which are functions of the instantaneous configuration of the stimulus' $n$ degrees of freedom. The numerical simulation in Section 3 and the Appendix is a specific example of this type of situation: namely, a stimulus with three degrees of freedom that is observed with sensors producing 20 measurements. In all cases like this, the Takens embedding theorem~\cite{Sauer} states that the mapping between the stimulus state $x$ and the multiplet of sensor signals is almost certainly invertible. Therefore, if a dimensional reduction technique~\cite{Tenenbaum} is used to define an
$n\mbox{-dimensional}$ $\tilde{x}$ coordinate system on the manifold of observed sensor signals, $x$ and $\tilde{x}$ will be related in an invertible fashion. In other words, because of the Takens' theorem, invertibility is almost guaranteed  as long as care is taken to equip the observing machine with a sufficient number of sensors. Notice that the invertibility of $f$ implies that the sensors of each observing machine define a ``measurement" coordinate system ($\tilde{x}$) on the stimulus state space, and, in fact, the only essential difference between machines equipped with different sensors is that they record the stimulus trajectory in different coordinate systems

Now consider the local second-order correlation matrix mentioned in Section~1 
\begin{equation}
g^{kl}(x)=<(\dot{x}_k-\bar{\dot{x}}_k) \, (\dot{x}_l-\bar{\dot{x}}_l)>_{x},
\end{equation}
and assume that this quantity approaches a definite limit as the neighborhood shrinks to zero around $x$. Because this correlation matrix transforms as a symmetric contravariant tensor, it can be taken to be a contravariant metric on the system's state space. Furthermore, as long as the local velocity distribution is not confined to a hyperplane in velocity space, this tensor is positive definite and can be inverted to form the corresponding covariant metric $g_{kl}$. Thus, under these conditions, the system's trajectory induces a non-singular metric on state space~\cite{Levin}.

How strong are the foregoing assumptions? The right side of Eq.(3) is expected to have a well-defined local limit if the trajectory densely covers a patch of state space and if its local distribution of velocities varies smoothly over that space.  Specifically, suppose that there is a density function $\rho(x,\dot{x})$, which varies smoothly with $x$ and which measures the fraction of total trajectory time that the trajectory spends in a small neighborhood $dx d\dot{x}$ of $(x, \dot{x}) \mbox{-space}$ (i.e., phase space). In that case, the limit in Eq.(3) certainly exists and is proportional to a second moment of that function. In Section 3, we show that the trajectories of a wide variety of classical physical systems are described by such density functions in phase space.

Our goal is to devise a way to test the data in any coordinate system (e.g., the $\tilde{x}$ coordinate system) in order to determine if it is separable. So, we will proceed by assuming that the data are separable, and then we will derive necessary conditions that the data-derived metric must satisfy in any coordinate system. First, let's assume that, in the $x$ coordinate system, the density function can be separated into the product of  two density functions $\rho(x,\dot{x}) = \rho_A(x_A,\dot{x_A})\rho_B(x_B,\dot{x_B})$, where $x_A$ and $x_B$ are components of $x$ with consecutive indices $x_{Ak} = x_k \mbox{ for } k = 1,2, \ldots ,n_A < n$ and $x_{Bk} = x_k \mbox{ for } k = n_A + 1,n_A + 2, \ldots ,n$. The factorizibility of the density function implies that the metric is block-diagonal in the $x$ coordinate system; i.e.,
\begin{equation}
g_{kl}(x)=
\left( \begin{array} 
{cc} 
g_A(x_A) & \mathbf{0} \\
\mathbf{0} & g_B(x_B) \\
\end{array} \right)_{kl}
\end{equation}
where $g_A$ and $g_B$ are $n_A \, \mbox{x} \, n_A$ and $n_B  \, \mbox{x} \, n_B$ matrices, $n_B = n - n_A$, and each $\mathbf{0}$ symbol denotes a null matrix of appropriate dimensions. 

Next, define the $A$ ($B$) subspace at each point $x$ to be the hyperplane through that point with constant $x_B$ ($x_A$). Projecting the trajectory's velocity vector at $x$ onto the $A$ and $B$ subspaces at that point separates it into components that represent the motion of the $A$ and $B$ processes, respectively. The operator that performs this projection onto the $A$ subspace is the $n \, \mbox{x} \, n$ matrix $A^k{}_l$
\begin{equation}
A^k{}_l =
\left( \begin{array} 
{cc} 
\mathbf{1} & \mathbf{0} \\
\mathbf{0} & \mathbf{0} \\
\end{array} \right)_{kl}
\end{equation}
where $\mathbf{1}$ is the $n_A \, \mbox{x} \, n_A$ identity matrix.  In other words, if $\dot{x}$ is the velocity of the stimulus at $x$, then $A^k{}_l \dot{x}_l$ is the velocity of the $A$ process, where we have used Einstein's convention of summing over repeated indices. The complementary projector onto the $B$ subspace is $B^k{}_{l} = \delta^k{}_{l} - A^k{}_{l}$, where $\delta^k{}_{l}$ is the Kronecker delta. In any other coordinate system (e.g., the $\tilde{x}$ coordinate system), the corresponding projectors ($\tilde{A}^k{}_l$ and $\tilde{B}^k{}_l$) are mixed-index tensor transformations of the projectors in the $x$ coordinate system; for example,
\begin{equation}
\tilde{A}^k{}_l(\tilde{x}) = \frac{\partial \tilde{x}_{k}}{\partial x_{k'}} \, \frac{\partial x_{l'}}{\partial \tilde{x}_l} \, A^{k'}{}_{l'}
\end{equation}.

Because the $A$ and $B$ projectors permit the local separation of the $A$ and $B$ processes, it will be useful to be able to construct them in the measurement ($\tilde{x}$) coordinate system. Our strategy for doing this is to find conditions that the projectors must satisfy in the $x$ coordinate system and then transfer those conditions to the $\tilde{x}$ coordinate system by writing them in coordinate-system-independent form. First, note that Eq.(5) implies that $A^k{}_l$ is idempotent
\begin{equation}
A^k{}_{k'} A^{k'}{}_{l} = A^k{}_l,
\end{equation}
and it is unequal to the identity and null matrices. Next, consider the Riemann-Christoffel curvature tensor of the stimulus state space~\cite{Weinberg} 
\begin{equation}
R^k{}_{lmn}(x) = - \frac{\partial \Gamma^k{}_{lm}}{\partial x_n} + \frac{\partial \Gamma^k{}_{ln}}{\partial x_m}
              + \Gamma^k{}_{im} \Gamma^i{}_{ln} - \Gamma^k{}_{in} \Gamma^i{}_{lm},
\end{equation}
where the affine connection $\Gamma^k_{lm}$ is defined in the usual way
\begin{equation}
\Gamma^k_{lm} = \frac{1}{2} g^{kn} (\frac{\partial g_{nl}}{\partial x_m} +
\frac{\partial g_{nm}}{\partial x_l} - \frac{\partial g_{lm}}{\partial x_n}).
\end{equation}
The block-diagonality of $g_{kl}$ in the $x$ coordinate system implies that $\Gamma^k_{lm}$ and $R^k{}_{lmn}$ are also block-diagonal in all of their indices. The block-diagonality of the curvature tensor, together with Eq.(5), implies
\begin{equation}
R^j{}_{klm}(x) A^k{}_{i} - A^j{}_{k} R^k{}_{ilm}(x) = 0
\end{equation}
at each point $x$. Covariant differentiation of Eq.(10) will produce other local conditions that are necessarily satisfied by separable data. It can be shown that these conditions are also linear constraints on the subspace projector because the projector's covariant derivative vanishes.

What is the intuitive meaning of Eq.(10)? Because of the block-diagonality of the affine connection in the $x$ coordinate system, it is easy to see that parallel transfer of a vector lying within the $A$ (or $B$) subspace at any point produces a vector within the $A$ (or $B$) subspace at the destination point. Consequently, parallel transfer of the corresponding projectors ($A^k{}_l$ and $B^k{}_l$) is path-independent. In particular, parallel transferring one of these projectors along the $i^{th}$ direction and then along the $j^{th}$ direction will give the same result as parallel transferring it along the $j^{th}$ direction and then along the $i^{th}$ direction. Equation (10) is the statement of this path-independent parallel transfer of the projectors that exist on separable manifolds. In contrast, for most inseparable Riemannian manifolds there are no non-trivial solutions of Eqs.(7) and (10). For example, on any intrinsically curved two-dimensional surface (e.g., a sphere), it is not possible to find a one-dimensional projector at each point (i.e., a direction at each point) that satisfies Eq.(10). This is because the parallel transfer of directions on such a surface is path dependent.

Notice that the quantities in Eqs.(7) and (10) transform as tensors when the coordinate system is changed. Therefore, these equations must be true \textit{in any coordinate system} on a separable state space. In particular, in the $\tilde{x}$ coordinate system that is defined by the sensors of the observing machine, we have
\begin{equation}
\tilde{A}^k{}_{k'}(\tilde{x}) \tilde{A}^{k'}{}_{l}(\tilde{x}) = \tilde{A}^k{}_l(\tilde{x})
\end{equation}
\begin{equation}
\tilde{R}^j{}_{klm}(\tilde{x}) \tilde{A}^k{}_i(\tilde{x}) - \tilde{A}^j{}_k(\tilde{x}) \tilde{R}^k{}_{ilm}(\tilde{x}) = 0.
\end{equation}
So far, we have shown that in any coordinate system on a separable space there must be non-trivial solutions to Eqs.(11) and (12); i.e., there are special projectors or directions at each point that parallel transfer path-independently. Thus, separability imposes a significant constraint on the curvature tensor of the space and, therefore, on the observed data. Likewise, if no solution of Eqs.(11) and (12) exists, we can immediately conclude that the data are not separable by any nonlinear transformation.

On the other hand, if the data are separable, we can use the solutions of Eqs.(11) and (12) to explicitly separate it; i.e., we can construct a transformation from the measurement coordinate system ($\tilde{x}$) to the source coordinate system ($x$). First, we solve these equations at a single point $\tilde{x}_0$ in order to find $\tilde{A}^k{}_{l}(\tilde{x}_0)$ and its complement $\tilde{B}^k{}_{l}(\tilde{x}_0)$. Then, the following procedure can be used to construct a geodesic coordinate system in which the metric is explicitly block-diagonal. First, select $n$ linearly independent small vectors $\delta{\tilde{y}}_{(i)}$ ($i = 1,2, \ldots ,n$) at $\tilde{x}_0$, and use $\tilde{A}^k{}_{l}(\tilde{x}_0)$ and $\tilde{B}^k{}_{l}(\tilde{x}_0)$ to project them onto the local $A$ and $B$ subspaces. Then, use the results to create a set of $n_A$ linearly independent vectors $\delta{\tilde{x}}_{(a)}$ ($a = 1,2, \ldots ,n_A$) and a set of $n_B$ linearly independent vectors $\delta{\tilde{x}}_{(b)}$ ($b = n_A + 1,n_A + 2, \ldots ,n$), which lie within the $A$ and $B$ subspaces, respectively. Finally: 1) starting at $\tilde{x}_0$, use the affine connection to repeatedly parallel transfer all $\delta{\tilde{x}}$ along $\delta{\tilde{x}}_{(1)}$; 2) starting at each point along the resulting geodesic path, repeatedly parallel transfer these vectors along $\delta{\tilde{x}}_{(2)}$;  ...  n) starting at each point along the most recently produced geodesic path, parallel transfer these vectors along $\delta{\tilde{x}}_{(n)}$. Each point in the neighborhood of $\tilde{x}_0$ is assigned the geodesic coordinate $s$ ($s_k, k = 1,2, \ldots ,n$), where each component $s_k$ represents the number of parallel transfers of the vector $\delta{\tilde{x}}_{(k)}$ that was required to reach it. If one visualizes these projection and parallel transfer procedures in the $x$ coordinate system of a separable space, it is not hard to see that the first $n_A$ components of $s$ (i.e., $s_A$)will be functions of $x_A$ and the last $n_B$ components of $s$ ($s_B$) will be functions of $x_B$. In other words, $s$ and $x$ will just differ by a coordinate transformation that is block-diagonal with respect to the subspaces.  Therefore, the metric will be block-diagonal in the $s$ coordinate system, just like it is in the $x$ coordinate system.  But, because $s$ is defined by coordinate-system-independent procedures, the same $s$ coordinate system will be constructed by performing these procedures in the measurement ($\tilde{x}$) coordinate system. In summary: separability necessarily implies that the subspace projectors satisfy Eqs.(11) and (12) at $\tilde{x}_0$  and that the metric will be block-diagonal in the geodesic ($s$) coordinate system computed from those projectors. 

We are now in a position to systematically determine if the observed data can be decomposed into independent subspaces. The first step is to use the observed measurements $\tilde{x}(t)$ to compute the metric (Eq.(3)), affine connection (Eq(9)), and curvature tensor (Eq.(8)) at one particular point $\tilde{x}_0$ in the state space. Next, Eqs.(11) and (12) are solved algebraically to find all possible subspace projectors $\tilde{A}^k{}_{l}(\tilde{x}_0)$ at that point. If non-trivial solutions are not found, we conclude that the observations are not separable into independent subspaces. If solutions are found, each one is used to construct an $s$ (geodesic) coordinate system. If the metric is not block-diagonal in the $s$ coordinate systems computed from any of these solutions, we conclude that the observations cannot be separated by any nonlinear transformation.

If the metric is block-diagonal in a geodesic coordinate system, we continue the separation procedure by applying all of the above procedures separately to the metrics on the $A$ and $B$ subspaces in order to see if they can be subdivided into smaller subspaces with independent second-order velocity statistics. This process is repeated until the metric is block-diagonal and has blocks that cannot be further subdivided in this way. The resulting geodesic coordinate systems comprise all coordinate systems in which the metric is block-diagonal, up to permutations of blocks and transformations of coordinates within blocks (or groups of blocks). Therefore, the coordinates of each independent source process must correspond to the blocks (or groups of blocks) of one of these geodesic coordinate systems. Additional criteria of statistical independence, such as those used in linear BSS~\cite{Hyvarinen}, can be employed to determine which blocks (or groups of blocks) of geodesic coordinates are truly independent (in the sense that they lead to factorization of the trajectory's density function). For example, we can check whether there is a vanishing second-order correlation between the components of any two blocks, we can determine if higher-order correlations among components from different blocks factorize into the products of lower-order correlations of components within blocks, etc. If any groups of blocks do not pass these tests, it may be necessary to aggregate them into larger blocks.

There is one exceptional situation that can arise in this serial decomposition procedure. If two or more one-dimensional subspaces are produced by repeatedly applying the second-order statistical criterion, each of the corresponding diagonal metric components (i.e., the corresponding \textit{1 x 1} blocks) can be transformed to unity by appropriate linear or nonlinear scale transformations on those subspaces. Therefore, these subspaces can be mixed by any rotation, without affecting the block-diagonality of the metric (i.e., without affecting the second-order statistical independence of $\dot{s}$). This unknown rotation could be determined by applying the above-mentioned criteria of statistical independence used in linear BSS~\cite{Hyvarinen}. Thus, in this exceptional case, the proposed methodology fails to completely define blocks of independent source variables, but it does reduce the nonlinear BSS problem to linear BSS. There is another way to understand the failure of Eqs.(11) and (12) to fully determine the source coordinate system for flat spaces. Because the curvature tensor vanishes on a flat space, Eq.(12) imposes no constraint on the subspace projectors in that case. This is related to the fact that, in a flat space, \textit{any} vectors or projectors at a given point can be parallel transferred to other locations in a path-independent manner, unlike a curved space in which only special projectors (or none at all) have this property. Therefore, in a flat space, we have more freedom (an undetermined rotation) in our choice of the projectors that can be used to construct a geodesic coordinate system in which the metric is diagonal.

\section*{\begin{center} 3. EXAMPLES \end{center}}

In this Section, we demonstrate large classes of stimulus trajectories, satisfying the assumptions in Section 2. In these cases: 1) the trajectory's statistics are described by a density function in phase space; 2) the trajectory-derived metric is well-defined and can be computed analytically; 3) there is a source coordinate system in which the density function is separable into the product of two density functions and in which the metric is block-diagonal. Many of these trajectories are constructed from the behavior of physical systems that could be realized in the laboratory. This Section concludes with a brief description of a numerical simulation of such a laboratory experiment. A detailed description of this simulation is given in the Appendix.

First, consider the energy of a physical process with $n$ degrees of freedom $x$ ($x_k \mbox{ for } k = 1,2, \ldots ,n$)
\begin{equation}
E(x,\dot{x}) = \frac{1}{2} \mu_{kl}(x) \dot{x}_k \dot{x}_l + V(x)
\end{equation}
where $\mu_{kl}$ and $V$ are some functions of $x$. Furthermore, suppose that
\begin{equation}
\mu_{kl}(x)=
\left( \begin{array} 
{cc} 
\mu_A(x_A) & \mathbf{0} \\
\mathbf{0} & \mu_B(x_B) \\
\end{array} \right)_{kl}
\end{equation}
\begin{equation}
V(x) = V_A(x_A) +  V_B(x_B)
\end{equation}
where $\mu_A$ and $\mu_B$ are $n_A \, \mbox{x} \, n_A$ and $n_B  \, \mbox{x} \, n_B$ matrices for $1 \leq n_A < n$ and $n_B = n - n_A$, where each $\mathbf{0}$ symbol denotes a null matrix of appropriate dimensions, and where $x_{Ak} = x_k \mbox{ for } k = 1,2, \ldots ,n_A$ and $x_{Bk} = x_k \mbox{ for } k = n_A + 1,n_A + 2, \ldots ,n$. These equations describe the degrees of freedom ($x_A$ and $x_B$) of almost any pair of classical physical systems, which do not exchange energy or interact with one another. A simple system of this kind consists of a particle with coordinates $x_A$ moving in a potential $V_A$ on a possibly warped two-dimensional frictionless surface with physical metric $\mu_{Akl}(x_A)$, together with a particle with coordinates $x_B$ moving in a potential $V_B$ on a two-dimensional frictionless surface with physical metric $\mu_{Bkl}(x_B)$. In the general case, suppose that the system intermittently exchanges energy with a thermal ``bath" at temperature T. This means that the system evolves along one trajectory from the Maxwell-Boltzmann distribution at that temperature and periodically jumps to another trajectory randomly chosen from that distribution. After a sufficient number of jumps, the amount of time the system will have spent in a small neighborhood $dxd\dot{x}$ of $(x,\dot{x})$ is given by the product of $dxd\dot{x}$ and a density function that is proportional to the Maxwell-Boltzmann distribution~\cite{Sears}
\begin{equation}
\mu(x)\exp{[-E(x,\dot{x})/kT]}
\end{equation}
where $k$ is the Boltzmann constant and $\mu$ is the determinant of $\mu_{kl}$. As described in Section 2, the existence of this density function means that a well-defined local velocity covariance matrix exists, and computation of Gaussian integrals shows that it is
\begin{equation}
<(\dot{x}_k-\bar{\dot{x}}_k) \, (\dot{x}_l-\bar{\dot{x}}_l)>_{x} = kT\mu^{kl}(x).
\end{equation}
where $\mu^{kl}$ is the contravariant tensor equal to the inverse of $\mu_{kl}$. It follows that the trajectory-induced metric on the stimulus state space is well-defined and is given by $g_{kl}(x) = \mu_{kl}(x)/kT$. Furthermore, Eq.(14) shows that the metric has a block-diagonal form in the $x$ coordinate system. This reflects the fact that the density function in Eq.(16) is the product of the density functions of the two non-interacting subsystems.

In order to test these ideas ``experimentally", we numerically simulated a stimulus with three degrees of freedom. Specifically, the stimulus consisted of two particles, one of which moved freely on a spherical surface and the other of which moved freely on a line. These particles were ``observed" by a simulated machine that was equipped with five pinhole cameras that suffered from nonlinear distortions of their optical paths. These sensors produced 20 numbers at each time point, consisting of the coordinates of the particles in the distorted images of all cameras. After dimensional reduction by locally linear embedding~\cite{Tenenbaum}, the measurement time series was blindly processed by the technique described in Section 2. Equations (11) and (12) were found to have non-trivial solutions corresponding to a two-dimensional subspace and a complementary one-dimensional subspace, and the metric was found to be nearly block-diagonal in the corresponding geodesic coordinate system. This geodesic coordinate system was expected to be identical to the separable coordinate system in which the system was originally defined, except for coordinate transformations confined to the individual subspaces. This was demonstrated by showing that the first two geodesic coordinate components correctly described an ``independent" two-dimensional process, which corresponded to the particle's path on the spherical surface. A detailed description of this ``experiment" is given in the Appendix. 

\vspace{20 mm}

\section*{\begin{center} 4. DISCUSSION \end{center}}

This paper outlines a new approach to nonlinear blind source separation, as well as nonlinear independent subspace separation. In many situations, the method solves the nonlinear BSS problem, or, at least, it reduces nonlinear BSS to linear BSS, without the use of probabilistic, parametric, or iterative procedures. The first step is to rephrase the problem in the following manner: 1) given a time series of observations in a sensor-defined coordinate system ($\tilde{x}$) on the stimulus state space, determine if there is another coordinate system (a source coordinate system $x$) in which groups of components are statistically independent of one another; 2) if such a coordinate system exists, find the transformation to it. The existence (or lack thereof) of such a source coordinate system is a coordinate-system-independent property of the stimulus' evolution (i.e., an intrinsic or invariant property). This is because, in \textit{all} coordinate systems, there either is or is not a transformation to such a source coordinate system. In general, differential geometry provides mathematical machinery for determining whether a manifold has a coordinate-system-independent property like this. In the case at hand, we can induce a geometric structure on the stimulus space by identifying its metric with the local second-order correlation matrix of the stimulus' velocity. Then, a necessary condition for BSS (the block-diagonalizibility of the metric everywhere) can be shown to impose constraints on the data-derived curvature tensor in all coordinate systems (including the measurement coordinate system). If the curvature tensor violates those conditions, the observations are not separable. However, if the curvature tensor satisfies those constraints, the $\tilde{x}$ metric can be used to construct a geodesic ($s$) coordinate system in which the metric has a block-diagonal form (i.e., in which groups of stimulus velocity components are statistically independent to second order). The coordinates of independent source processes must correspond to blocks (or groups of blocks) of these geodesic coordinates. An exceptional situation arises if there is a multidimensional flat subspace. In that case, the desired source coordinate system may differ from a geodesic coordinate system by a rotation on this subspace. Additional statistical criteria, such as the second-order and higher-order statistical constraints used in linear BSS~\cite{Hyvarinen}, can be used to compute the unknown rotation and to determine if any blocks of geodesic coordinates need to be fused into larger blocks.

In a sense, the methodology in this paper is an application of the general framework described in an earlier report~\cite{Levin}. That paper introduced the idea of using the local velocity correlation matrix as a metric on the stimulus state space. The parallel transfer operation, derived from that metric, was then employed to describe relative stimulus locations. As an example of such a description, suppose that stimuli \textit{A, B,} and \textit{C} differ by small stimulus transformations, and suppose that a more distant stimulus $D$ can be described as being related to \textit{A, B,} and \textit{C} by the following procedure: ``$D$ is the stimulus that is produced by the following sequence of operations: 1) start with stimulus $A$ and parallel transfer the vectors $A \to B$ and $A \to C$ along $A \to B$ 23 times; 2) start at the end of the resulting geodesic and parallel transfer $A \to C$ along itself 72 times". All such statements about relative stimulus locations derived from parallel transfer are coordinate-system-independent. They are also machine-independent because the only essential difference between machines equipped with different sensors is that they record the stimulus state in different coordinate systems (Section 2). Figures 3(c) and 3(d) (see Appendix) demonstrate an example of the machine-independence of such statements about relative stimulus locations. Specifically, this figure shows how many parallel transfers of the vectors $\delta \tilde{x}_{i}$ were required to reach each point on the test pattern, as computed by a machine $\tilde{O}b$ equipped with five pinhole camera sensors (narrow black lines) and as computed by a different machine $Ob$ that directly sensed the values of $x$ (thick gray lines). In Ref.~\cite{Levin}, it was emphasized that different machines can use this technology to navigate through stimulus space and to represent stimuli in the same way, even though they do not communicate with one another. The entire set of such statements about relative stimulus locations constitutes a rich stimulus representation that is intrinsic to the stimulus' evolution in the sense that it does not depend on extrinsic factors such as the observer's choice of a coordinate system in which the stimulus is viewed (i.e., the observer's choice of sensors). The current paper shows how a ``blinded" observer can glean another intrinsic property of the stimulus' evolution, namely its separability.

What are the limitations on the application of this method? As discussed in Section 2, the metric is expected to be well-defined if the trajectory densely covers a patch of state space and if its local distribution of velocities varies smoothly over that space. In any event, the metric certainly exists if the trajectory is described by a density function in phase space. In Section 3, we showed that these conditions were satisfied by trajectories describing a wide variety of physical systems. In practical applications, one must have observations that cover state space densely enough in order to compute the metric, as well as its first and second derivatives (required to compute the affine connection and curvature tensor).  In the numerical simulation in Section 3 and the Appendix, approximately 8.3 million short trajectory segments (containing a total of 56 million points) were used to compute the metric and curvature tensor on a \textit{32 x 32 x 32} grid on the three-dimensional state space. Of course, if the dimensionality of the state space is higher, even more data will be needed. So, like a human infant, machines of this type must observe stimuli for relatively long periods of time in order to be able to discern their separable nature. If a machine's sensors are changing in time, the method should be run in an adaptive mode, in which the metric is derived from recently observed data. Of course, the adaptation time must be long enough to allow the stimulus trajectory to cover state space with the required density. There are few other limitations on the applicability of the technique. For example, it can be applied to machines equipped with just about any sensors, as long as each sensor's output is an instantaneous function of the stimulus state and as long as more than $2n$ sensors are used to observe an $n \mbox{-dimensional}$ stimulus. In particular, because the method is blinded to the physical nature of each sensor, it can effortlessly fuse the outputs of sensors from different modalities. Furthermore, computational expense is not prohibitive. The computation of the metric is the most CPU-intensive part of the method. However, it can be distributed over multiple processors by dividing the observed data into ``chunks" corresponding to different time intervals, each of which is sent to a different processor where its contribution to the metric is computed. As additional data is accumulated, it can be processed separately and then added into the time average of the data that was used to compute the earlier estimate of the metric (Eq.(3)). Thus, the earlier data need not be processed again, and only the latest observations need to be kept in memory.

It is interesting to return to the biological phenomena that have inspired work on BSS, as mentioned in Section 1.  Many psychological experiments suggest that human perception is remarkably sensor-independent. Specifically, suppose that an individual's visual sensors are changed by having the subject wear goggles that distort and/or invert the observed scene. Then, after a sufficiently long period of adaptation, most subjects perceive the world in approximately the same way as they did before the experiment~\cite{Held}. An equally remarkable phenomenon is the approximate universality of human perception: i.e., the fact that perceptions seem to be shared by individuals with different sensors (e.g., different ocular anatomy and different microscopic brain anatomy), as long as they have been exposed to similar stimuli in the past. Thus, many human perceptions seem to represent properties that are ``intrinsic" to stimuli in the sense that they do not depend on the way the stimuli are observed (i.e., they don't depend on the type of sensors or on the nature of the sensor-defined coordinate system on state space). This paper and an earlier one~\cite{Levin} describe a method of finding such ``inner" properties of a sufficiently dense stimulus trajectory. Is it possible that the human brain somehow extracts these particular invariants from sensory data?  The only way to test this speculation is to perform biological experiments to determine if the human brain actually utilizes the specific metric and geometric structure described in these two papers.

\section*{\begin{center} ACKNOWLEDGMENTS \end{center}}

The author is indebted to Michael A. Levin for insightful comments and for a critical reading of the manuscript. This work was performed with the partial support of the National Institute of Deafness and Communicative Disorders.
 
\section*{\begin{center} APPENDIX: NUMERICAL EXAMPLE \end{center}}

In this Appendix, the scenario described in Section 3 is illustrated by the numerical simulation of a stimulus with three degrees of freedom. The stimulus was comprised of two moving particles of unit mass, one moving on a transparent frictionless curved surface and the other moving on a frictionless line. Figure 1 shows the curved surface, which consisted of all points within one radian of a randomly chosen point.
\begin{figure}[|tbp]
\centering
\includegraphics[width=2.8in]{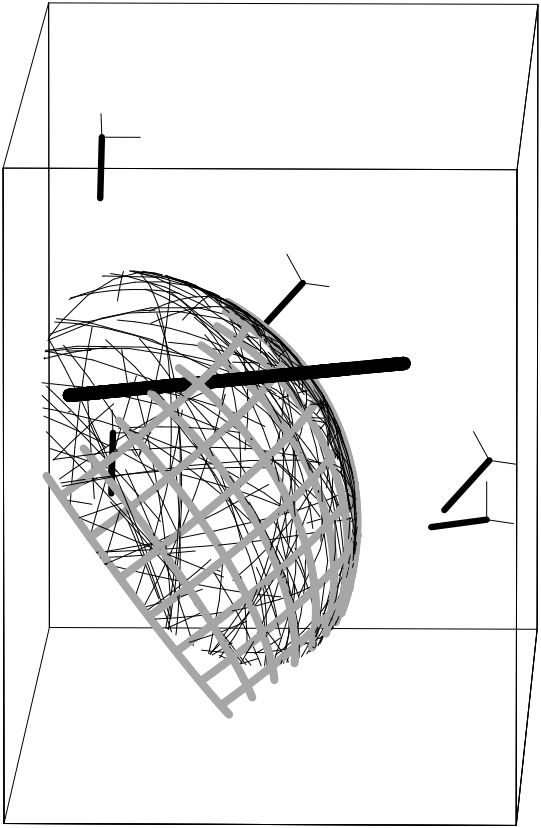}
\caption{\small{The thin black curved lines comprise a small sample of the trajectory segments traversed by the particle that was confined to a spherical surface, and the long thick black line shows the corresponding trajectory segments of the second particle constrained to a straight line. The stimulus was ``watched" by five simulated pinhole cameras. Each small triplet of orthogonal straight lines shows the relative position and orientation of a camera, with the long thick line of each triplet being the perpendicular to a camera focal plane that was represented by the two short thin lines of each triplet. One camera is nearly obscured by the spherical surface. The thick gray curved lines show some latitudes and longitudes on the spherical surface.}}
\end{figure}
Figure 1 also shows that the curved surface and line were oriented at arbitrarily-chosen angles with respect to the simulated laboratory coordinate system. Both particles moved freely, and they were in thermal equilibrium with a bath for which $kT = 0.01$ in the chosen units of mass, length, and time. As in Section 3, the source trajectory was created by temporally concatenating approximately 8.3 million short trajectory segments randomly chosen from the corresponding Maxwell-Boltzmann distribution, given by Eqs.(13-16) with $\mu_A$ equal to the metric of the spherical surface, $\mu_B$ equal to a constant, and  $V_A = V_B = 0$. Figure 1 shows a small sample of those trajectory segments.

The particles were ``watched" by a simulated machine $\tilde{O}b$ equipped with five pinhole cameras, which had arbitrarily chosen positions and faced the cylinder/line with arbitrarily chosen orientations (Fig. 1). The image created by each camera was transformed by an arbitrarily chosen second-order polynomial, which varied from camera to camera. In other words, each pinhole camera image was distorted by translational shift, rotation, rescaling, skew, and quadratic deformations that simulated the effect of a distorted optical path between the particles and the camera's focal plane. The output of each camera was comprised of the four numbers representing the two particles' locations in the distorted image on its ``focal" plane. As the particles moved, the cameras created a time series of sensor multiplets, each of which consisted of the 20 numbers produced by all five cameras at one time point. Figure 2(a) shows the first three principal components of the system's trajectory through the corresponding 20-dimensional space.
\begin{figure}[|tbp]
\centering
\includegraphics[width=3.4in]{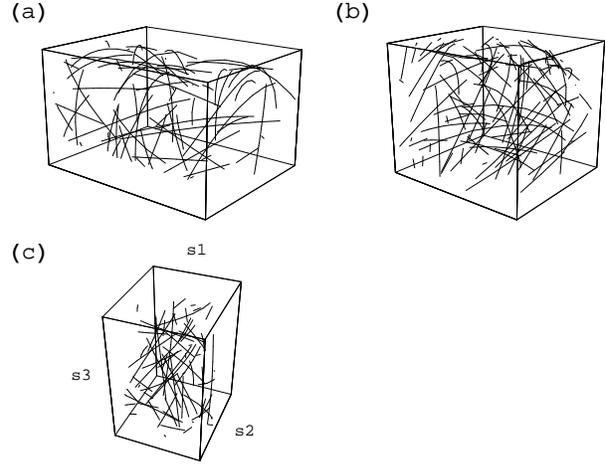}
\caption{\small{(a) A small sample of stimulus trajectory segments after they were mapped into the 20-dimensional space of camera outputs. Only the first three principal components are shown. (b) A small sample of stimulus trajectory segments after dimensional reduction was used to map them from the 20-dimensional space onto the three-dimensional measurement ($\tilde{x}$) space. (c) A small sample of stimulus trajectory segments after they were transformed from the $\tilde{x}$ coordinate system to the geodesic ($s$) coordinate system (i.e., the ``experimentally" determined source coordinate system).}}
\end{figure}
A dimensional reduction technique (locally linear embedding~\cite{Tenenbaum}) was applied to the full 20-dimensional time series in order to identify the underlying three-dimensional measurement space and to establish a coordinate system ($\tilde{x}$) on it. The value of $\tilde{x}$ associated with each stimulus state (i.e., each location of the two particles) defined the sensor state (or measurement state) of $\tilde{O}b$. Because the stimulus state space had dimensionality $n = 3$ and because the sensor multiplets had more than $2n$ components, the Takens embedding theorem~\cite{Sauer} virtually guaranteed that there was a one-to-one mapping between the stimulus and measurement states. Therefore, the $\tilde{x}$ coordinate system, defined by dimensional reduction of the sensor data, was a coordinate system on the stimulus state space as well. The exact nature of that coordinate system depended on the the channels and sensors used to make the machine's measurements (e.g., it depended on the positions, orientations, and optical path distortions of the five pinhole cameras). Figure 2(b) shows typical trajectory segments in the $\tilde{x}$ coordinate system.

Next, Eqs.(3), (9), and (8) were used to compute the metric, affine connection, and curvature tensor in this coordinate system. Then, Eqs.(11) and (12) were solved at a point $\tilde{x}_0$. One pair of solutions was found, representing a local projector onto a two-dimensional subspace and the complementary projector onto a one-dimensional subspace. Following the procedure in Section 2, we selected three small linearly independent vectors $\delta \tilde{y}_{(i)}$ ($i = 1,2,3$) at $\tilde{x}_0$, and we used the projectors at that point to project them onto the putative $A$ and $B$ subspaces. Then, the resulting projections were used to create a set of two linearly independent vectors $\delta{\tilde{x}}_{(a)}$ ($a = 1, 2$) and a single vector $\delta{\tilde{x}}_{(3)}$ within the $A$ and $B$ subspaces, respectively. Finally, the geodesic ($s$) coordinate system was constructed by using the affine connection to parallel transfer these vectors throughout the neighborhood of $\tilde{x}_0$. After the metric was transformed into the $s$ coordinate system, it was found to have a nearly block-diagonal form, consisting of a \textit{2 x 2} block and a \textit{1 x 1} block. In other words, the time derivatives of the corresponding groups of $s$ variables were approximately statistically independent to second-order, as expected. Because the two-dimensional subspace had non-zero intrinsic curvature (proportional to the intrinsic curvature of the underlying spherical surface), it could not be decomposed into smaller (i.e., one-dimensional) independent subspaces. Therefore, in this example, the data were separable, and the source coordinate system was the geodesic ($s$) coordinate system, which was unique up to coordinate transformations on each block and up to subspace permutations.

In order to demonstrate the accuracy of the separation process, we defined ``test lines" that had known projections onto the independent subspaces used to define the stimulus. Then, we compared those projections with the test pattern's projection onto the independent subspaces that were ``experimentally" determined by the proposed method. First, we defined a Cartesian $x$ coordinate system in which $x_A$ was the position (longitude, latitude) of the particle on the spherical surface and in which $x_B$ was the position of the other particle along the line (Fig. 1). In this coordinate system, the test lines consisted of straight lines that were oriented at various angles with respect to the $x_B = 0$ plane and that projected onto the grid-like array of latitudes and longitudes in that plane. In other words, each line corresponded to a path generated by moving the first particle along a latitude or longitude of the sphere and simultaneously moving the second particle along its constraining line. The points along these test lines were ``observed" by the five pinhole cameras to produce corresponding lines in the 20-dimensional space of the cameras' output (Fig. 3(a)).
\begin{figure}[|tbp]
\centering
\includegraphics[width=3.4in]{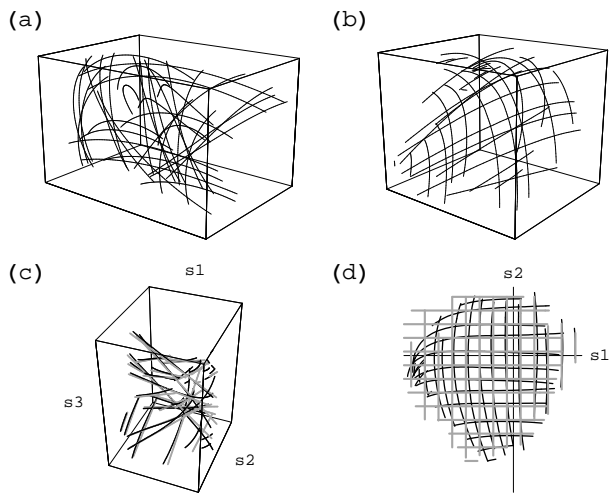}
\caption{\small{(a) The test lines after they were mapped into the 20-dimensional space of camera outputs. The figure only depicts the resulting pattern's projection onto the space of the first three principal components of the 20-dimensional stimulus trajectory. (b) The test lines after  dimensional reduction was used to map them from the 20-dimensional space onto the three-dimensional measurement ($\tilde{x}$) space traversed by the trajectory segments. (c) The thin black lines show the test pattern after it was transformed from the $\tilde{x}$ coordinate system to the geodesic ($s$) coordinate system, which comprises the ``experimentally" derived source coordinate system. The thick gray lines show the test lines in the comparable exact source coordinate system. (d) The thin and thick black lines show the first two components of the test patterns in (c). These collections of lines represent the projection of the test pattern onto the ``experimentally" derived two-dimensional independent subspace and onto the exactly known independent subspace, respectively.}}
\end{figure}
These lines were then mapped onto lines in the $\tilde{x}$ coordinate system by means of the same procedure used to dimensionally reduce the trajectory data (Fig. 3(b)). Finally, the test pattern was transformed from the $\tilde{x}$ coordinate system to the $s$ coordinate system, the geodesic coordinate system that comprised the ``experimentally" derived source coordinate system. As mentioned above, the $s$ coordinate system was the only possible separable coordinate system, except for arbitrary coordinate transformations on each subspace. Therefore, it should be the same as the $x$ coordinate system (an exactly known source coordinate system), except for such block-diagonal transformations. The nature of that coordinate transformation depended on the choice of vectors that were parallel transfered to define the geodesic ($s$) coordinate system on each subspace. In order to compare the test pattern in the ``experimentally" derived source coordinate system ($s$) with the appearance of the test pattern in the exactly known source coordinate system ($x$), we picked $\tilde{x}_0$ and the $\delta \tilde{y}$ vectors so that the $s$ and $x$ coordinate systems would be the same, as long as the independent subspaces were correctly identified by $\tilde{O}b$. Specifically: 1) $\tilde{x}_0$ was chosen to be the mapping of the origin of the $x$ coordinate system, which was located on the sphere's equator and at the line's center; 2) $\delta{\tilde{y}}_{(1)}$ and $\delta{\tilde{y}}_{(2)}$ were chosen to be mappings of vectors projecting along the equator and the longitude, respectively, at that point; 3) all three $\delta \tilde{x}$ were normalized with respect to the metric in the same way as the corresponding unit vectors in the $x$ coordinate system. Figure 3(c) shows that the test pattern in the ``experimentally" derived source coordinate system consisted of nearly straight lines (narrow black lines), which almost coincided with the test pattern in the exactly known source coordinate system (thick gray lines). Figure 3(d) shows that the test pattern projected onto a grid-like pattern of lines on the ``experimentally" determined $A$ subspace (narrow black lines), and these lines nearly coincided with the test pattern's projection onto the exactly known $A$ subspace (thick gray lines). These results indicate that the proposed BSS method correctly determined the source coordinate system. In other words, the ``blind" observer $\tilde{O}b$ was able to separate the state space into two independent subspaces, which were nearly the same as the independent subspaces used to define the stimulus.

\end{document}